# Combat Models for RTS Games

Alberto Uriarte and Santiago Ontañón

*Abstract*—Game tree search algorithms, such as Monte Carlo Tree Search (MCTS), require access to a forward model (or "simulator") of the game at hand. However, in some games such forward model is not readily available. This paper presents three forward models for two-player attrition games, which we call "combat models", and show how they can be used to simulate combat in RTS games. We also show how these combat models can be learned from replay data. We use STARCRAFT as our application domain. We report experiments comparing our combat models predicting a combat output and their impact when used for tactical decisions during a real game.

## I. INTRODUCTION

A significant number of different artificial intelligence (AI) algorithms that play Real-Time Strategy (RTS) games, like Monte Carlo Tree Search (MCTS) [1], assume the existence of a forward model that allows to advance (or predict) the state after executing a certain action in the current game state. While this assumption is reasonable in certain domains, such as Chess or Go where simulating the effect of actions is trivial, precise descriptions of the effect of actions are not available in some domains. Thus, techniques for defining forward models are of key importance to RTS game AI, since they would allow the application of game tree search algorithms, such as MCTS.

In this paper we focus on the problem of defining "forward models" for RTS games, limiting ourselves to only simulate combat situations. We propose to model a combat as an *attrition game* [2] (an abstract combat game where individual units cannot move and only their damage and hit points are considered). We then propose three models (*TS-Lanchester$^2$*, *Sustained* and *Decreasing*) to simulate the evolution of an attrition game over time, and compare them with the different existing models in the literature. Finally, we identify the set of parameters that these models require, and study how to automatically learn those parameters from replay data. We argue that while forward models might not be available, logs of previous games might be available, from where the result of applying specific actions in certain situations can be observed. For example, consider the STARCRAFT game (used as our testbed), where precise definitions of the effects of unit actions is not available, but large collections of replays are available.

In the rest of this paper, we will use the term "combat model" to refer to a "forward model" in the context of a combat simulation. Moreover, notice that forward models can be used as evaluation functions (simulating the combat to the end, and then assessing the final state), and thus, we will also study the use of combat models as evaluation functions.

The remainder of this paper is organized as follows. First, we provide background on combat models in RTS games and their applications. Then, we define a combat as an attrition game and we propose new models for RTS. After that, we explain the parameters needed for these combat models and how they can be learned. Then, we describe how to use these combat models within an MCTS framework for RTS games. Finally, we present an empirical evaluation in STARCRAFT.

## II. BACKGROUND

Real-Time Strategy (RTS) games in general, and STAR-CRAFT in particular, have emerged as a fruitful testbed for new AI algorithms [3], [4]. One of the most recurrent techniques for tactical decisions are those based on game tree search, like *alpha-beta* search [5] or MCTS [6]–[9]. Game tree search algorithms require some representation of the game state, a forward model that gives us the game state resulting from applying an action to another game state, and an evaluation function that assigns reward scores to game states.

Concerning the representation of the game state, using a direct representation of the game state in RTS games is not feasible with the current search algorithms due the resulting large branching factor (number of possible actions a player can perform in a given game state). For example, the branching factor in STARCRAFT can reach numbers between $30^{50}$ and $30^{200}$ [4]. To palliate this problem several approaches have been explored. For example, Chung et al. [10] applied Monte Carlo planning to an RTS game by simplifying the decision space: assuming that each player can choose only one amongst a finite set of predefined plans. Balla and Fern [6] performed an abstraction of the game state representation grouping the units in groups but keeping information of each individual unit at the same time, and allowing only two types of actions per group: attack and merge with another group. Kovarsky and Buro [11] considered all units as non movable units and without attack range, i.e., a unit is able to attack any other unit at any time. Churchill and Buro [7], simplified the possible actions to a set of scripted behaviors to reduce the search with their proposed *Portfolio Greedy Search* algorithm; Justesen et al. [9] extended Churchill and Buro's work allowing different units to perform different scripted behavior and investigating how to cluster the units that may perform the same script. Finally, Uriarte and Ontañón [8] used an abstraction based on dividing the terrain in regions using the BroodWar Terrain Analysis (BWTA[1]), and grouped the units by type and region.

Alberto Uriarte and Santiago Ontañón are with the Computer Science Department, Drexel University, Philadelphia, PA 30309 USA (e-mail: santi@cs.drexel.edu; albertouri@cs.drexel.edu).

[1]https://bitbucket.org/auriarte/bwta2

## III. COMBAT IN RTS GAMES

One of the mechanics present in almost all RTS games is combat. In a combat situation each player commands an army of units to defeat the opponent. We will model combats by using *attrition games* [2]. An *attrition game* is a simultaneous-move two-player game in a directed graph, where each node has health, attack power and belongs to a player. A directed edge denotes that node $x$ can deal damage to node $y$ and when the health of a node reaches to zero the node is removed from the graph. The goal of the game is to destroy all opposing nodes. At each round each node can select a target and at the end of the round all the damage is done simultaneously. This abstraction can be seen as a target selection problem since no movement is involved in the process and nodes do not have complex properties. A similar combat abstraction was studied by Kovarsky and Buro [11] in the context of heuristic search, and they showed that finding the optimal strategy in these games is PSPACE-hard [2].

Using *attrition games* as a starting point, we define a combat as a tuple $C = \langle A, B, E \rangle$, where:
- $A$ and $B$ are the sets of *units* (or nodes) of each player,
- a *unit* is a tuple $u = \langle h, ap \rangle$, where $h$ is the health of the unit and $ap$ its attack power,
- $E$ is the set of directed edges between units (there is an edge from a unit $a$ to a unit $b$, when $a$ can attack $b$).

In the particular case of STARCRAFT, a unit is extended to be a tuple $u = \langle type, pos, hp, s, e \rangle$, where:
- $type$ is the unit type, which defines some common properties like maximum Hit Points (HP), weapon damage, or weapon "cooldown" (time between attacks),
- $pos = \langle x, y \rangle$ is the unit position (used to determine which units are inside the attack range of other units),
- $hp$ are the hit points, $s$ is the shield, and $e$ is the energy.

The *unit type* and *energy* are used to set the node's attack power; and the *unit type*, *hit points* and *shield* are used to define the node's health. Finally, a *combat model* $m$ is a function that given an initial (0) combat sate, it predicts the final ($f$) combat state and the length of the combat ($t$): $m : \langle A_0, B_0, E \rangle \rightarrow \langle A_f, B_f, t \rangle$.

## IV. EXISTING COMBAT MODELS FOR STARCRAFT

Different types of combat models have been proposed in the literature for RTS games. In this section, we will first categorize those existing models into three broad classes: *low-level final state prediction models* (which try to model the combat as close as possible to the real game); *high-level final state prediction models* (which use some level of abstraction but try to give an accurate prediction of the remaining army composition at the end of a combat); and *high-level winner prediction models* (which only predict the winner).

### A. Low-level Final State Prediction Models

The most representative low-level model for STARCRAFT is SPARCRAFT[2]. SPARCRAFT was developed via a large human effort observing the behavior of different STARCRAFT units frame by frame. Despite being extremely accurate for small combat scenarios, SPARCRAFT is not exhaustive (for example, it does not model unit collisions, and it does not support some unit types such as spell casters or transports) due to the tremendous amount of effort that it would take to model the complete STARCRAFT game. However, there has been recent work [12] on improving the accuracy of the unit movement simulation in SPARCRAFT. Despite reaching a high degree of fidelity, a perfect low-level model is impossible to achieve due some stochastic components of STARCRAFT. For example, each unit has a waiting time between attacks (a.k.a. "cooldown") that can be increased randomly, or it has a chance of success to hit a target depending on the environment (units in a high ground against units in a low ground).

[2]https://github.com/davechurchill/ualbertabot/tree/master/SparCraft

The biggest handicap of low-level models like SPARCRAFT is computational cost. First, simulating a combat requires simulating all the individual details of the game state. Also, if these models are used as the representation for game tree search, the branching factor of the resulting search tree can be very high. Branching factor at the level of abstraction at which SPARCRAFT operates can be as high as $10^{200}$ [8]. Another problem is that these models do not generalize well (i.e., SPARCRAFT cannot be used to simulate WARCRAFT).

However, the advantages are having a high-fidelity simulation of the effect of actions/scripts, and making mapping the state of the actual game to the state that the model needs easy.

### B. High-level Final State Prediction Models

In order to reduce computational cost, combat models over game state abstractions have been proposed. For example, Soemers [13] proposed a combat model based on Lanchester's Square Law. Lanchester's Laws [14] assume that combats are between two homogeneous armies; a *combat effectiveness* (or Lanchester attrition-rate coefficient) for each army is computable; and any unit in an army can attack any opponent unit. However, this is not the case in most RTS games where, for example, not all units can attack flying units. To overcome these shortcomings, several extensions have been proposed [15], but only some of them have been explored in the context of RTS games. Soemers computed the combat effectiveness of heterogeneous armies aggregating the mean DPF (Damage per Frame) of an army, divided by the mean HP of the opponent's army. Similarly, Stanescu et al. [16] used the Lanchester's Generalized Law (a generalization of the Square Law), but this time learning the combat effectiveness from replay data. As a consequence of only considering homogeneous armies, Lanchester's Laws can predict the size of the surviving army at the end of a combat, but not which specific units survived (since armies are assumed to be homogeneous). Stanescu et al. only used this model to predict the winner of a combat, and thus did not address this problem. Soemers addressed this by determining which units survived the combat in a heterogeneous army by selecting the remaining units randomly (while matching with the predicted army size).

The advantages of high-level state prediction models are that they offer a faster simulation time than low-level models, at the price of losing some low-level details due to abstraction.

*C. High-level Winner Prediction Models*

Finally, some approaches focus directly on just predicting the winner of a combat, instead of providing a prediction of the final state of the combat. These models can be classified in two subcategories: heuristic and machine learning-based.

Concerning *heuristic models*, Uriarte [17] defined a basic heuristic assuming that both armies continuously deal their starting amount of DPF to each other, until one of the armies is destroyed. Kovarsky and Buro [11] proposed a function that gives more importance to having multiple units with less HP than only one unit with full HP: Life Time Damage 2 (LTD2).

$$LTD2 = \sum_{u \in A} \sqrt{u.HP} \times u.DPF - \sum_{u \in B} \sqrt{u.HP} \times u.DPF$$

Nguyen et al. [18] considered that RTS games properties are non-additive, i.e., if $\mu(X)$ is the effectiveness by a unit combinations $X$, $\mu(X_1 \cup X_2) > \mu(X_1) + \mu(X_2)$ or $\mu(X_1 \cup X_2) < \mu(X_1) + \mu(X_2)$; and they proposed to use fuzzy integrals to model this aggregation problem.

Concerning *machine learning-based models*, Synnaeve and Bessière [19] clustered armies based on their unit compositions and they were able to predict the winner using a Gaussian mixture model; Stanescu et al. [20] improved the previous work defining a Bayesian network and using a Gaussian Density Filtering to learn some features of the model. Finally, Sánchez-Ruiz [21] experimented with some machine learning algorithms (such as LDA, QDA, SVM or kNN) to predict the winner of small combats over time.

The key advantage of these models is their low computational cost, while being able to be used as *evaluation functions*.

V. PROPOSED COMBAT MODELS FOR STARCRAFT

This section presents three new high-level final state prediction models aiming at predicting the final state of a combat with higher accuracy than existing models. Additionally, in order to be useful in the context of game tree search, these models can: predict the final state of a combat (which units survived, even for heterogeneous army compositions), predict the duration of a combat, and simulate partial combats (i.e., not only fights-to-the-end).

*A. Target-Selection Lanchester's Square Law Model (TS-Lanchester$^2$)*

*TS-Lanchester$^2$* is related to the model proposed by Soemers [13] in that they are both based on Lanchester's Square Law [14], but differs in two key aspects: First, *TS-Lanchester$^2$* is based on the original formulation of Lanchester's Square Law (as formulated in [15]), while Soemers uses a reformulation in order to support army reinforcements. Second, and most important, our model incorporates a target selection policy, used at the end of the simulation to determine exactly which units remain at the end of the combat.

Since we use the original formulation of the Square Law instead of the formulation used by Soemers, we include the exact version of the equations used by our model here. Specifically, we model a combat as:

$$\frac{d|A_t|}{dt} = -\alpha |B_t|, \frac{d|B_t|}{dt} = -\beta |A_t|$$

where $|A_t|$ denotes the number of units of army $A$ at time $t$, $|B_t|$ is the analogous for army $B$, $\alpha$ is the number of $A$'s units killed per time unit by $B$'s units (a.k.a *combat effectiveness or Lanchester attrition-rate coefficient*) and $\beta$ is the analogous for army $B$. The intensity of the combat is defined as $I = \sqrt{\alpha\beta}$.

Given an initial combat state $\langle A_0, B_0 \rangle$, where armies belong to each player, the model proceeds as follows:

1) First the model computes the *combat effectiveness* of an heterogeneous army for each army as:

$$\alpha = \frac{avgDPF(B,A)}{\overline{HP}(A)}, \beta = \frac{avgDPF(A,B)}{\overline{HP}(B)}$$

where $\overline{HP}(A)$ is the average HP of units in army $A$ and $avgDPF(B,A)$ is the average DPF that units in army $B$ can deal to units in army $A$ computed as:

$$\overline{DPF}_{air}(B) \times \frac{\overline{HP}_{air}(A)}{\overline{HP}_{air}(A) + \overline{HP}_{ground}(A)} +$$
$$\overline{DPF}_{ground}(B) \times \frac{\overline{HP}_{ground}(A)}{\overline{HP}_{air}(A) + \overline{HP}_{ground}(A)}$$

2) The winner is then predicted:

$$\frac{|A_0|}{|B_0|} \begin{cases} > R_\alpha & A \text{ wins} \\ = R_\alpha & \text{draw} \\ < R_\alpha & B \text{ wins} \end{cases}$$

where $|A_0|$ and $|B_0|$ are the number of initial units in army $A$ and $B$ respectively, and $R_\alpha$ is the relative effectiveness $\left(R_\alpha = \sqrt{\frac{\alpha}{\beta}}, R_\beta = \sqrt{\frac{\beta}{\alpha}}\right)$.

3) And the length of the combat is computed as:

$$t = \begin{cases} \frac{1}{2I} ln\left(\frac{1 + \frac{|B_0|}{|A_0|} R_\alpha}{1 - \frac{|B_0|}{|A_0|} R_\alpha}\right) & \text{if } A \text{ wins} \\ \frac{1}{2I} ln\left(\frac{1 + \frac{|A_0|}{|B_0|} R_\beta}{1 - \frac{|A_0|}{|B_0|} R_\beta}\right) & \text{if } B \text{ wins} \end{cases}$$

Notice, however, that since this is a continuous model, the special case where there is a tie (both armies annihilate each other) is problematic, and results in a predicted time $t = \infty$. In our implementation, we detect this special case, and use our *Sustained* model (Section V-B) to make a time prediction instead.

4) Once the length is known, the remaining units are:

$$\begin{cases} |A| = \sqrt{|A_0|^2 - \frac{\alpha}{\beta}|B_0|^2} & |B| = 0 \quad \text{if } A \text{ wins} \\ |B| = \sqrt{|B_0|^2 - \frac{\beta}{\alpha}|A_0|^2} & |A| = 0 \quad \text{if } B \text{ wins} \end{cases}$$

To determine which units survived, the model use a *target selection policy*, which determines in what order the units are removed. The final combat state $\langle A_f, B_f, t \rangle$ is defined as the units that were not destroyed at time $t$.

This model predicts the final state given a time $t$ as follows:

$$|A_t| = \frac{1}{2}\left((|A_0| - R_\alpha |B_0|)e^{It} + (|A_0| + R_\alpha |B_0|)e^{-It}\right)$$

$$|B_t| = \frac{1}{2}\left((|B_0| - R_\beta|A_0|)e^{It} + (|B_0| + R_\beta|A_0|)e^{-It}\right)$$

*TS-Lanchester*$^2$ has two input parameters: a vector of length $k$ (where $k$ is the number of different unit types, in STARCRAFT $k = 163$) with the $DPF$ of each unit type; and a *target selection policy*. In Section VI, we propose different ways in which these input parameters can be generated.

### B. Sustained DPF Model (Sustained)

$Sustained$ is an extension of the model presented by Uriarte in [17]. It assumes that the amount of damage an army can deal does not decrease over time during the combat (this is obviously an oversimplification, since armies might lose units during a combat, thus decreasing their damage dealing capability) but it models which units can attack each other with a greater level of detail than *TS-Lanchester*$^2$.

Given an initial combat state $\langle A_0, B_0 \rangle$, where armies belong to each player, the model proceeds as follows:

1) First, the model computes how much time each army needs to destroy the other. In some RTS games, such as STARCRAFT, units might have a different DPF (*damage per frame*) when attacking to different types of units (e.g., air vs ground units), and some units might not even be able to attack certain other units (e.g., walking swordsmen cannot attack a flying dragon). Thus, for a given army, we can compute its $DPF_{air}$ (the aggregated DPF of units with a weapon that can attack air units), $DPF_{ground}$ (the aggregated DPF of units with a ground weapon) and $DPF_{both}$ (aggregated DPF of the units with both weapons, ground and air). After that, the model computes the time required to destroy all air and ground units separately:

$$t_{air}(A,B) = \frac{HP_{air}(A)}{DPF_{air}(B)}$$

$$t_{ground}(A,B) = \frac{HP_{ground}(A)}{DPF_{ground}(B)}$$

where $HP(A)$ is the sum of the hit points of all the units. Then, the model computes which type of units (air or ground) would take longer to destroy, and $DPF_{both}$ is assigned to that type. For instance, if the air units take more time to kill (i.e., $t_{air}(A,B) > t_{ground}(A,B)$) we recalculate $t_{air}(A,B)$ as:

$$t_{air}(A,B) = \frac{HP_{air}(A)}{DPF_{air}(B) + DPF_{both}(B)}$$

These equations are symmetric, therefore $t_{air}(B,A)$ is calculated analogously to $t_{air}(A,B)$. And finally the global time to kill the other army is computed:

$$t(A,B) = max(t_{air}(A,B), t_{ground}(A,B))$$

2) The combat length time $t$ is computed as:

$$t = min(t(A_0,B_0), t(B_0,A_0))$$

3) After that, the model computes which units does each army have time to destroy of the other army in time $t$. For this purpose, this model takes as input a *target*

---

**Algorithm 1** Decreasing.

1: **function** DECREASING($A, B, DPF, targetSelection$)
2:    SORT($A, targetSelection$)
3:    SORT($B, targetSelection$)
4:    $i \leftarrow 0$                                    ▷ index for army A
5:    $j \leftarrow 0$                                      ▷ index for army B
6:    **while true do**
7:       $t_b \leftarrow$ TIMETOKILLUNIT($B[j], A, DPF$)
8:       $t_a \leftarrow$ TIMETOKILLUNIT($A[i], B, DPF$)
9:       **while** $t_b = \infty$ **and** $j < B.size$ **do**
10:         $j \leftarrow j + 1$
11:         $t_b \leftarrow$ TIMETOKILLUNIT($B[j], A, DPF$)
12:       **while** $t_a = \infty$ **and** $i < A.size$ **do**
13:         $i \leftarrow i + 1$
14:         $t_a \leftarrow$ TIMETOKILLUNIT($A[i], B, DPF$)
15:       **if** $t_b = \infty$ **and** $t_f = \infty$ **then break**
16:       **if** $t_b = t_a$ **then**                    ▷ draw
17:         $A$.ERASE($i$)
18:         $B$.ERASE($j$)
19:       **else**
20:         **if** $t_b < t_a$ **then**            ▷ A wins
21:            $A[i].HP \leftarrow A[i].HP - $ DPF$(B) \times t_b$
22:            $B$.ERASE($j$)
23:         **else**                             ▷ B wins
24:            $B[j].HP \leftarrow B[j].HP - $ DPF$(F) \times t_a$
25:            $A$.ERASE($i$)
26:       **if** $i >= A.size$ **or** $j >= B.size$ **then break**
27:    **return** $A, B$

---

*selection policy*, which determines the order in which an army will attack the units in the groups of the other army. The final combat state $\langle A_f, B_f, t \rangle$ is defined as all the units that were not destroyed at time $t$.

$Sustained$ has two input parameters: a vector of length $k$ with the $DPF$ of each unit type; and a *target selection policy*. Moreover, compared to *TS-Lanchester*$^2$, $Sustained$ takes into account that not all units can attack all other units (e.g., ground versus air units). On the other hand, it does not take into account that as the combat progresses, units die, and thus the DPF of each army decreases, which is precisely what the model presented in the following subsection aims to address.

### C. Decreasing DPF Model (Decreasing)

$Decreasing$ is more fine grained than $Sustained$, and considers that when a unit is destroyed, the DPF that an army can deal is reduced. Thus, instead of computing how much time it will take to destroy the other army, it only computes how much time it will take to kill one unit, selected by the *target selection policy*. The process is detailed in Algorithm 1, where first the model determines which is the next unit that each player will attack using the *target selection policy* (lines 2-5); after that it computes the expected time to kill the selected unit using TIMETOKILLUNIT($u, A, DPF$) (where $u$ is the unit that army $A$ will try to kill, and $DPF$ is a matrix specifying the DPF that each unit type can deal to each other unit type); the target that should be killed first is eliminated from the combat state, and the HP of the survivors is updated (lines 16-25). The model keeps doing this until one army is completely annihilated or it cannot kill more units.

*Decreasing* has two input parameters: a $k \times k$ matrix $DPF$ (where $k$ is the number of different unit types), where $DPF_{i,j}$ is the DPF that a unit of type $i$ would deal to a unit of type $j$; and a *target selection policy*. Let us now focus on how these parameters can be acquired for a given RTS game.

## VI. COMBAT MODELS PARAMETERS

All three proposed models take as input the unit's DPF (Damage per Frame) and a *target selection policy*. This sections shows how these parameters can be calculated. For each of the two parameters, we performed experiments by learning these parameters from replay data (assuming no information about the game rules), and also using a collection of baseline configurations for them (e.g., taking the DPF values directly from the StarCraft manual).

- **Effective DPF**. In practice, the DPF a unit can deal does not just depend on the damage and cool-down of their weapon, but is also affected by their maneuverability, special abilities, the specific enemy being targeted, and the skill of the player. Therefore, we call *effective DPF* to the expected DPF that a unit can actually deal in a real game situation. We will compute effective DPF at two different levels of granularity:
  - Per unit: a vector of length $k$ (the number of unit types) with the effective DPF of each unit type.
  - Per match-up: a $k \times k$ matrix $DPF$, where $DPF(i,j)$ represents the DPF that a unit of type $i$ is expected to deal to a unit of type $j$.

  Moreover, the $DPF(i)$ vector can be derived from the $DPF(i,j)$ matrix as follows:

  $$DPF(i) = \min_j DPF(i,j)$$

  Thus, below we will focus on just how the effective DPF matrix can be learned from game replays. For comparison purposes, we will compare this against the **static** DPF ($u.damage/u.cooldown$) calculated directly from the STARCRAFT unit stats.

- **Target selection policy**. This is a function that given a set of units $A$, determines which unit $u \in A$ to attack next. We will also show how this can be learned from replays, and will compare the learned policy against two baselines: **random**, which selects units randomly from a uniform distribution; and maximizing the **destroy score**. In STARCRAFT the unit's destroy score is approximately $2 \times u.mineralCost + 4 \times u.gasCost$ (although some units a higher or lower score, based on their special abilities).

### A. Learning Combat Parameters

This section presents how to apply an off-line learning method from game replays to learn the combat parameters. Given a collection of replays, they can be preprocessed in order to generate a training set consisting of all the *combats* that occur in these replays. The construction of the specific dataset we used in our experiments is explained on Section VIII. Given a training set consisting of a set of combats, the combat model parameters can be learned as follows:

- **Effective DPF** matrix. For each combat in the dataset, the following steps are performed:
  1) First, for each player $p$, count the number of units that can attack ground units ($n^p_{ground}$), air units ($n^p_{air}$) or both ($n^p_{both}$).
  2) Then, let $K$ be a set containing a record $(t_i, u_i)$ for each unit destroyed during the combat, where $t_i$ is the frame where unit $u_i$ was destroyed. For each $(t_i, u_i) \in K$, the total damage that had to be dealt up to $t_i$ to destroy $u_i$ is: $d_i = u_i.HP + u_i.shield$ where *HP* and *shield* are the hit points and shield of the unit at the start of the combat. We estimate how much of this damage was dealt by each of the enemy units, by distributing it uniformly among all the enemies that could have attacked $u_i$. For instance, if $u_i$ is an air unit, the damage is split as:

  $$d_i^{split} = \frac{d_i}{n^p_{air} + n^p_{both}}$$

  After that, for each unit $u$ that could attack $u_i$, two global counters are updated:

  $$damageToType(u, u_i) += d_i^{split}$$

  $$timeAttackingType(u, u_i) += t_{i-1} - t_i$$

  where $t_{i-1}$ is the time at which the previous unit of player $p$ was destroyed (or 0, if $u_i$ was the first).
  3) After all the combats in the dataset have been processed, the effective DPF matrix is estimated as:

  $$DPF(i,j) = \frac{damageToType(i,j)}{timeAttackingType(i,j)}$$

- **Target selection policy**. We used a **Borda Count** method [22] to estimate the target selection policy used by the players in our dataset. The idea is to iterate over all the combats in the training set and each time a type of unit is killed for the first time, give that type $n-i$ points where $n$ is the number of different unit types in the army and $i$ the order the units were killed. For example, if an army is fighting against marines, tanks and workers ($n = 3$) and if the first units to be destroyed are tanks, then marines and then workers, the assigned scores will be: 2 points for the tank, 1 point for the marines and 0 points for the workers. After analyzing all the combats the average Borda Count of each unit type is computed and this is the score to sort the targets in order of preference. We repeated this estimation three times (generating three different target selection policies): one for the case when the attacking army is only composed by ground units, another one for armies with only air units and the last one for mixed armies. We observed, as expected, that the attacking armies composed by only ground units, prioritize enemies with ground weapons; and vice-versa for armies with only air units.

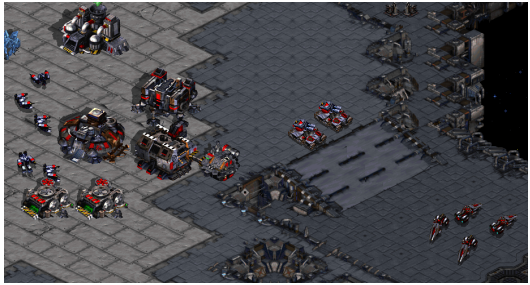

Fig. 1. Snapshot of a STARCRAFT game.

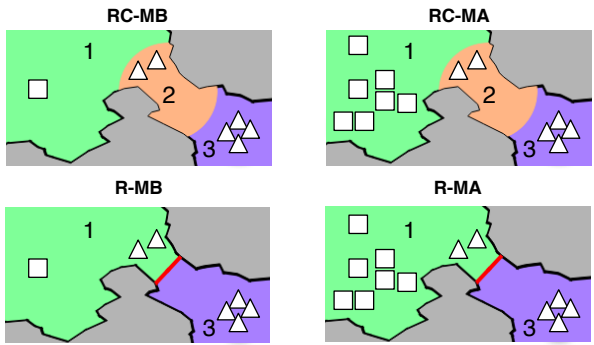

Fig. 2. Representation of a game state using different high-level abstraction with the ID of each region. Triangles are military units. Squares are buildings.

## VII. COMBAT MODELS IN GAME TREE SEARCH

In this section we will show how combat models can be integrated in to a MCTS framework to play RTS games, and STARCRAFT in particular. In order to deal with the enormous branching factor occurring in RTS games, we abstract both the game state and action set. Moreover, we only consider military unit movement and attacking as part of this MCTS framework, all other tasks (resource gathering, unit training, etc.) are not considered as part of the search. We incorporate this MCTS approach in a STARCRAFT playing bot taking charge of all the military units. In the next subsections we first describe the high-level game state representation we use; then we describe the set of high-level actions considered in our framework; and finally, how this is integrated into an actual RTS-playing bot (how a low-level state is mapped to a high-level state, and then how high-level actions are mapped down to low-level actions that can be executed in the game).

### A. High-Level State Representation

We propose to represent the map of an RTS game as a graph where each node corresponds to a region of the map, and edges represent adjacent regions. Additionally, instead of modeling each individual unit, we will group all the units of the same type inside each region into groups. We will also assume that units inside the same region are close enough so that they can attack each other, and units in different regions are far enough so they cannot attack each other.

*1) Map decomposition:* Given a STARCRAFT map, we decompose it into a graph of regions using Perkins' algorithm [23] (implemented in the BWTA library). This results in a set of regions, where each region is associated with a specific set of tiles in the low-level map of the game. Since the map is invariant, this process only has to be done once, at the beginning of the game.

TABLE I
GROUPED UNITS OF PLAYER 1 USING ABSTRACTION RC-MB.

| Plyr. | Type | Size | Avg. HP | Rgn. | Action | Target | End |
|---|---|---|---|---|---|---|---|
| 1 | Base | 1 | 1500 | 1 | N/A | - | - |
| 1 | Tank | 2 | 150 | 2 | Move | 3 | 230 |
| 1 | Vulture | 4 | 80 | 3 | Idle | - | 400 |

Moreover, regions found by Perkins' algorithm are separated by chokepoints (narrow passages). Since most of RTS combats happen in these locations, we can conceivably expand the graph returned by Perkins' algorithm by turning each chokepoint into a region (this idea was also explored by Synnaeve and Bessière [24]). We compared the performance of our approach both with and without adding these additional chokepoint regions. Chokepoint regions are added in the following way: given the output of Perkins' algorithm, and given a set of chokepoints $C$, we create a new region for each chokepoint $c \in C$ containing all the tiles in the map whose distance to the center of $c$ is smaller than the radius of the chokepoint (as computed by Perkins' algorithm). The graph of regions is then updated accordingly.

*2) Group of units:* The units are grouped by unit type and region. For each group, the following information is stored: *Player* (owner of the group), *Type* (type of units in the group), *Size* (number of units), *Average HP* (average hit points of all units in the group), *Region* (which region is this group in), *Action* (which action is this group currently performing), *Target* (the target region of the action, if applicable) and *End* (in which game frame is the action estimated to finish).

*3) Variations:* We experimented with four variations of the proposed high-level state representation:

- **R-MB**: Graph of regions is the output of Perkins' algorithm. All military units (except workers) are considered in the abstraction. The only structures considered are the bases (like *Command Centers* or *Nexus*).
- **R-MA**: Like R-MB, but including all structures.
- **RC-MB**: Like R-MB, but adding chokepoints as additional regions.
- **RC-MA**: Like RC-MB, but including all structures.

Figure 1 shows a screenshot of a situation in a STARCRAFT game, and Figure 2 graphically illustrates how this would be represented with the four different abstractions defined previously. The actual internal representation of the high-level game state is a matrix with one row per unit group (Table I).

### B. High-Level Actions

We define the following set of possible actions for each high-level group: *N/A*, *Move*, *Attack* and *Idle*:

- *N/A*: only for buildings as they cannot perform any action,
- *Move*: move to an adjacent region,
- *Attack*: attack enemies in the current region, and
- *Idle*: do nothing during 400 frames.

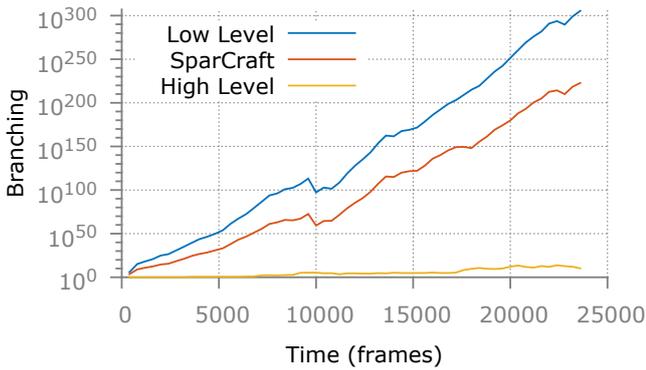

Fig. 3. Comparison of the branching factor using our high-level representation with respect to the branching factor using the low-level STARCRAFT game states and SPARCRAFT abstraction.

For example, the groups of player 1 in Table I using a RC-MB abstraction can execute the following actions:
- Bases: *N/A*.
- Tanks: Assuming that the Tanks were not executing any action, they could *Move* to region 1 or 3, or stay *Idle*. They cannot *Attack* since there is no enemy in region 2.
- Vultures: Assuming that the Vultures were not executing any action, they can *Move* to region 2, or stay *Idle*.

Since player 1 can issue any combination of those actions to her units, the branching factor of this example would be $(1) \times (2+1) \times (1+1) = 6$. To compare the branching factor resulting from this high-level representation with that of the low-level representation, we analyzed the branching factor in STARCRAFT game states every 400 frames during a regular game (Figure 3): *Low Level* is the actual branching factor of STARCRAFT when all actions of all units are considered in the low-level game state representation, SPARCRAFT is the branching factor of the low-level game state only considering combat units, and *High Level*, is the branching factor using our high-level representation. As expected, the branching factor using the low-level representation of the game state is very large (reaching values of up to $10^{300}$ in our experiment). Our high-level abstraction is able to keep the branching factor relatively low (reaching a peak of about $10^{10}$).

### C. Connecting Low-Level and High-Level States and Actions

In order to incorporate our MCTS approach into an actual STARCRAFT playing bot, we need to define a mapping between low-level states and high-level states. Moreover, our search process does not produce the exact low-level actions that the combat units need to perform, but high-level actions such as moving from one region to another, or attacking a particular region. Thus, we assume that there is a low-level agent that translates the low-level state to our high-level representation and then is capable of translating the actions defined above into actual low-level actions.

Most STARCRAFT bots are decomposed in several individual agents that perform different tasks in the game, such as scouting or construction [4]. One of such agents is typically in charge of combat units, and is in charge of controlling a military hierarchy architecture. The high-level game-tree search approach presented in this paper is designed to replace such agent. As described above, our MCTS approach assigns actions to groups of units. Our STARCRAFT bot uses the intermediate concept of *squads* to control groups of units, which is analogous to the concept of groups in the high-level representation. Unfortunately, generating the set of groups in a given low-level game state given the set of squads is not trivial, since, for example, units in a given squad might appear to split in two groups when moving from one region to another, as some units arrive earlier than others, thus breaking squad continuity. Additionally, if two squads with units of the same type arrive to the same region, they would be grouped together into a single large group according to the high-level representation, which might not be desirable.

To address these problems, given a squad $q$ in the low-level, we add each unit $u \in q$ to the high-level state, recording which high-level group the unit is being mapped to. Once all the units in the squad have been added to the high-level state, all the units are reassigned to the high-level group to which most of the units in $q$ had been assigned. The *abstract layer* records the set of squads that were mapped to each of the high-level groups (notice that a high-level group can have more than one squad since two different squads in the same region with the same unit type will be mapped to the same group).

When simulating a combat using any of our combat models during the execution of MCTS, some units of the game state need to be removed from the simulation (since our combat models assume all units can attack and can be destroyed). First, the *invincible* and *harmless* units (i.e., the units that cannot receive or deal damage given the current armies composition) are removed from the simulation (but not from the game state). Then, the combat simulation is split in two: first we simulate the combat using only military units and after that, we simulate the combat of the winner against the *harmless* units that were removed initially from the simulation, in order to have an accurate estimate of the length of the combat.

Once the low-level state has been mapped to a high-level game state, we can use a game-tree search algorithm to get an action for each group (which can be mapped to orders for each squad for execution by the other agents of the STARCRAFT playing bot). Since RTS games are real-time, we perform a search process periodically (every 400 game frames). After each iteration, the order of each squad is updated with the result of the search. The specific configuration of MCTS used in our different experiments is described in the experimental evaluation section. The remainder of this paper describes our experimental evaluation in STARCRAFT. We start by describing how exactly we obtained the training set to train our combat models, and then we present experimental results of their performance.

## VIII. EXTRACTING COMBATS FROM GAME REPLAYS

To be able to learn the parameters required by the combat models from replays, we need a dataset of combats. Other researchers achieved this by creating artificial datasets making the default STARCRAFT AI to play against itself [21] and recording the state in traces; or using a crafted combat model (SPARCRAFT) to run thousands of combat situations [16], [20], while others used data from professional human players [19], [25], [26]. In this work we use the latter, to capture the performance of each unit as commanded by an expert.

### A. Combat Records

Let us define a *combat record* as a tuple $CR = \langle t_0, t_f, R, A_0, B_0, A_f, B_f, K, P \rangle$, where:

- $t_0$ is the initial frame when the combat started and $t_f$ the final frame when it finished,
- $R$ is the reason why the combat finished. The options are:
  - **Army destroyed**. One of the two armies was totally destroyed during the combat.
  - **Peace**. None of the units were attacking for the last $x$ frames. In our experiments $x = 144$, which is 6 seconds of game play.
  - **Reinforcement**. New units participating in the battle. This happens when units, that were far from the battle when it started, begin to participate in the combat.
  - **Game end**. This occurs when a combat is still going on, but one of the two players surrender the game.
- $A_0$ and $B_0$ are the armies of each player at the start of the combat, and $A_f$ and $B_f$ are the armies of each player at the end of the combat.
- $K = \{(t_1, u_1), \ldots, (t_n, u_n)\}$ where $t_i$ is the time when unit $u_i$ was killed.
- $P$ is the list of passive unites, i.e., units that did not participated in the combat, like workers keep mining while being attacked.

Given this definition, let us now introduce our approach to identify combats in STARCRAFT replays.

### B. Detecting Combat Start

One of the hardest part is how to define *when* a combat starts and ends. In Synnaeve and Bessière's work [19] they consider a new combat when a unit is killed. Although this helps to prune all non combat activities involved in an RTS game, in some cases we are losing useful information. For example, imagine the situation when a ranged unit is kiting a melee unit and the melee unit is killed at the end. If we start tracking the combat after the kill, it will look like the ranged unit killed the melee unit without effort, when in fact it took a while and a smart move-and-hit behavior. Hence, we start tracking a new combat if a *military unit* is *aggressive* or *exposed* and not already in a combat:

- A *military unit* is a unit that can deal damage (by a weapon or a magic spell), detect cloaked units, or a transporter.
- A unit is *aggressive* when it has the order to attack or is inside a transport.
- A unit is *exposed* if it is in the attack range of an *aggressive* enemy unit.

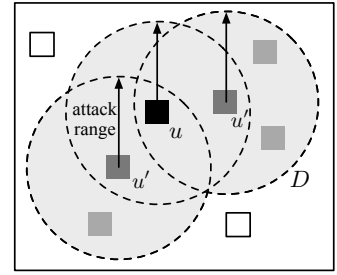

Fig. 4. Black filled square ($u$) triggers a new combat. Only filled squares are added to the combat tracking, i.e., the union of all $inRange(u')$.

### C. Units Involved in Combat

Any military unit $u$, that is either aggressive or exposed and not already in a combat, will trigger a new combat. To know which units are part of the combat triggered by $u$, let us define $inRange(u)$ to be the set of units in attack range of a unit $u$. Now, let $D = \cup_{u' \in inRange(u)} inRange(u')$. $A_i$ is the subset of units of $D$ belonging to player $A$ at time $t_i$; and $B_i$ is the subset of units of $D$ belonging to player $B$ at time $t_i$. Figure 4 shows a representation of a combat triggered by unit $u$ (the black filled square) and the units in the combat (the filled squares). Notice that a *military unit* is considered to take part in a combat even if at the end it does not participate, but it will be marked as *passive*.

### D. Combat Dataset

Now that we have a definition of a *combat record*, a dataset can be generated directly by processing replays. STARCRAFT replays of professional player games are widely available, and several authors have compiled collections of such replays in previous work [19], [25]. Since STARCRAFT replays only contain the mouse click events of the players (this is the minimum amount of information needed to reproduce the game in STARCRAFT), we do not know the full game state at a given point (no information about the location of the units or their health). Thus, we need to run the replay in STARCRAFT and record the required information using BWAPI[3]. This is not a trivial task since if we record the game state at each frame, we might have too much information (some consecutive recorded game state variables might have the same value) and the size of the data captured can grow too much to be processed efficiently. Some researchers proposed to capture different information at different resolutions to have a good trade-off of information resolution. For example, Synnaeve and Bessière [19] proposed recording information at three different levels of abstraction:

- General data: all BWAPI events (like unit creation, destruction or discovery). Economical situation every 25 frames and attack information every frame. It uses the heuristic explained above to detect attacks.
- Order data: all orders to units. It is the same information we will get parsing the replay file outside BWAPI.
- Location data: the position of each unit every 100 frames.

[3]https://github.com/bwapi/bwapi

On the other hand, Robertson and Watson [27] proposed a uniformed information gathering, recording all the information every 24 frames or every 7 frames during attacks to have a better resolution than the previous work.

In our case we only need the combat information, so we used the analyzer from Synnaeve and Bessière, but with our proposed heuristic for detecting combats (described above) [4].

## IX. Experimental Evaluation

In order to evaluate the performance of each combat model we first collected a dataset of combats from game replays (Section IX-A); we compared then the winner prediction accuracy over the combat dataset (Section IX-B); after that we evaluated the accuracy of predicting the final state (Section IX-C). Then, in order to observe the performance in a real game, we incorporated the combats models as *forward models* in a MCTS algorithm appropriate for RTS games (MCTSCD [8]). In the context of MCTSCD, we evaluated the gameplay performance achieved with different abstractions (Section IX-D) and different combat models (Section IX-E).

### A. Combat Dataset

We extracted the combats from 600 replays (100 for each race matchup: Terran versus Terran, Terran versus Protos, etc.). This resulted in a dataset of 140,797 combats where:
- 99,598 combats ended by *reinforcement*,
- 40,820 combats ended in *peace*,
- 11,454 combats ended with one *army destroyed*,
- 653 combats ended by *game end*.

To evaluate the performance of the combat models, we only used the subset of combats that ended with one *army destroyed*. To form the training set, we also removed combats with Vulture's mines (to avoid problems with friendly damage) and combats where an army was passive (none of their units fought back). This resulted in a dataset of 6,066 combats with an average combat length of 169.68 frames (max 3,176, min 1, combats with tiny durations are because of 1-shot-kills), an average of 9.4 units per combat (max 93, min 2) and an average of 2.95 different types of units per combat (max 10, min 2). Unfortunately, SPARCRAFT does not support all types of units, therefore we also generated a subset of 4,558 combats that are valid for SPARCRAFT in order to compare results against SPARCRAFT. When using learned parameters (DPS learn or Borda Count target selection), we report the result of a 10-fold cross validation.

### B. Winner Prediction Accuracy

In this first experiment we assessed the winner prediction accuracy of each combat model. When comparing against SPARCRAFT, we simulate the combat in SPARCRAFT by using one of the predefined scripted behaviors (*Attack-Closest (AC)*, *Attack-Weakest (AW)*, *Attack-Value (AV)*, *Kiter-Closest (KC)*, *Kiter-Value (KV)*, and *No-OverKill-Attack-Value (NOK-AV)*). For our proposed models (*TS-Lanchester*$^2$, *Sustained* and *Decreasing*) we experimented with two DPF matrices: *static* and *learn* as described in Section VI. When predicting the winner, the target selection policy only affects to the *Decreasing* model, therefore we experimented with all three target selection policies (Random, Destroy Score and Borda Count) only for the *Decreasing* model. Finally, we also compare against evaluating the initial combat state with the LTD and LTD2 evaluation functions, and predicting the winner based on the resulting evaluation. Since LTD and LTD2 use the unit's DPF, we tested the models with both DPF matrices (static and learn). Notice that LTD and LTD2 use the actual unit's HP from the low-level game state (like SPARCRAFT), in contrast of *TS-Lanchester*$^2$, *Sustained* and *Decreasing* use the average HP of each group of units.

[4]Source code of our replay analyzer, along with our dataset extracted from replays, can be found at https://bitbucket.org/auriarte/bwrepdump

TABLE II
WINNER PREDICTION ACCURACY OF EACH COMBAT MODEL

| Combat Model | Full Dataset | SPARCRAFT Dataset |
|---|---|---|
| SPARCRAFT (AC) | N/A | 88.02% |
| SPARCRAFT (AW) | N/A | 87.76% |
| SPARCRAFT (AV) | N/A | 87.76% |
| SPARCRAFT (NOK-AV) | N/A | 88.02% |
| SPARCRAFT (KC) | N/A | 84.69% |
| SPARCRAFT (KV) | N/A | 84.82% |
| LTD static | 90.37% | 91.82% |
| LTD learn | 82.05% | 86.55% |
| LTD2 static | 92.09% | 93.22% |
| LTD2 learn | 83.66% | 83.70% |
| *TS-Lanchester*$^2$ static | **94.10**% | **94.45**% |
| *TS-Lanchester*$^2$ learn | 90.66% | 91.33% |
| *Sustained* static | 93.60% | 94.12% |
| *Sustained* learn | 90.38% | 91.05% |
| Target Selection Policy: Random | | |
| *Decreasing* static | 93.55% | 93.92% |
| *Decreasing* learn | 91.10% | 92.32% |
| Target Selection Policy: Destroy Score | | |
| *Decreasing* static | 93.57% | 93.79% |
| *Decreasing* learn | 91.14% | 92.21% |
| Target Selection Policy: Borda Count | | |
| *Decreasing* static | 89.72% | 93.77% |
| *Decreasing* learn | 91.21% | 92.21% |

Table II shows the winner prediction accuracy of each combat model in the full dataset (6,066 combats) and in the SPARCRAFT compatible dataset (4,558 combats). Concerning the three combat models, we can see that *TS-Lanchester*$^2$ tends to obtain better results than *Sustained* or *Decreasing*. We can also see that the static DPF matrix tends to result in higher performance than the learned DPF matrix (although prediction accuracy is still very high with the learned matrix, showing it is a viable alternative for domains for which DPF of units is not known in advance). Concerning target selection policies, since this is a winner prediction task, they only play a role on the *Decreasing* model, and we see that except for one anomaly (*Decreasing* static with Borda count), they all perform similarly. Overall, the configuration that achieved best results is *TS-Lanchester*$^2$ static.

All our models achieved a higher winner prediction accuracy than SPARCRAFT. Our hypothesis is that SPARCRAFT achieves a lower accuracy because it simulates a combat by making players follow a set of predefined scripts, which might differ from the way in which human players control the units. Concerning the performance of LTD and LTD2, we see that they perform well (but below the performance of our model) only when using the static DPF matrix.

TABLE III
FINAL STATE AVERAGE SIMILARITY OF EACH COMBAT MODEL, AND TIME
(IN SECONDS) TO SIMULATE ALL THE COMBATS IN OUR DATASET.

| Combat Model | Full Dataset | SPARCRAFT Dataset | |
|---|---|---|---|
| | Similarity | Similarity | Time |
| SPARCRAFT (AC) | N/A | 0.8652 | 0.2328 |
| SPARCRAFT (AW) | N/A | 0.8550 | 0.2084 |
| SPARCRAFT (AV) | N/A | 0.8564 | 0.2218 |
| SPARCRAFT (NOK-AV) | N/A | 0.8592 | 0.1871 |
| SPARCRAFT (KC) | N/A | 0.8556 | 0.7495 |
| SPARCRAFT (KV) | N/A | 0.8522 | 0.7673 |
| Target Selection Policy: Random | | | |
| $TS\text{-}Lanchester^2$ static | 0.9088 | 0.9154 | 0.0096 |
| $TS\text{-}Lanchester^2$ learn | 0.8767 | 0.8853 | 0.0070 |
| $Sustained$ static | 0.9014 | 0.9093 | 0.0063 |
| $Sustained$ learn | 0.8660 | 0.8771 | 0.0061 |
| $Decreasing$ static | 0.9028 | 0.9103 | 0.0055 |
| $Decreasing$ learn | 0.8892 | 0.8939 | 0.0060 |
| Target Selection Policy: Destroy Score | | | |
| $TS\text{-}Lanchester^2$ static | 0.9056 | 0.9142 | 0.0072 |
| $TS\text{-}Lanchester^2$ learn | 0.8726 | 0.8840 | 0.0068 |
| $Sustained$ static | 0.9016 | 0.9111 | 0.0076 |
| $Sustained$ learn | 0.8679 | 0.8801 | 0.0062 |
| $Decreasing$ static | 0.9031 | 0.9113 | 0.0058 |
| $Decreasing$ learn | 0.8858 | 0.8937 | 0.0054 |
| Target Selection Policy: Borda Count | | | |
| $TS\text{-}Lanchester^2$ static | **0.9106** | **0.9166** | 0.0069 |
| $TS\text{-}Lanchester^2$ learn | 0.8766 | 0.8866 | 0.0067 |
| $Sustained$ static | 0.9016 | 0.9111 | 0.0065 |
| $Sustained$ learn | 0.8656 | 0.8772 | 0.0061 |
| $Decreasing$ static | 0.8987 | 0.9119 | 0.0056 |
| $Decreasing$ learn | 0.8893 | 0.8950 | 0.0053 |

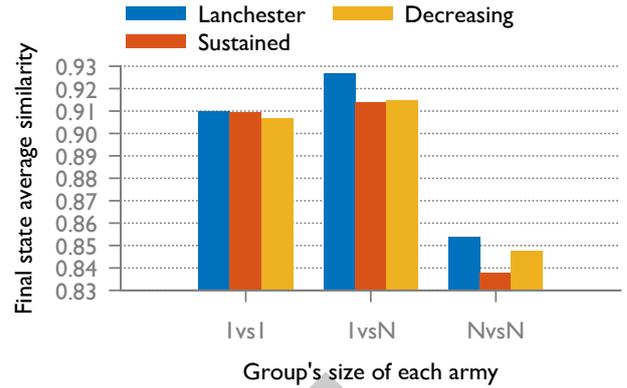

Fig. 5. Final state average similarity of each model using a Borda Count target selection policy and clustered by number of groups on each army.

### C. Final State Prediction Accuracy

This section compares the combat models based on how well do they predict the final game state of the combat (instead of just the winner). We compared against SPARCRAFT with the same configurations as before; and our proposed models (*TS-Lanchester*$^2$, $Sustained$ and $Decreasing$) with both DPF matrices (static and learn) and all three target selection policies (Random, Destroy Score and Borda Count) as described in Section VI.

In order to assess the similarity of the predicted state by our combat models and the actual end state in the training set, we used a measure inspired in the Jaccard index (a well known similarity measure between sets: the size of their intersection divided by the size of their union). Given a combat in the training set with initial state $S = A_0 \cup B_0$, and final state $F = A_f \cup B_f$, and the final state prediction generated by a combat model $F' = A'_f \cup B'_f$, the similarity between $F$ and $F'$ is defined as:

$$similarity(S, F', F) = 1 - \frac{||S \cap F| - |S \cap F'||}{|S|}$$

Table III shows the average similarity of the predictions generated by the different combat models with respect to the actual outcome of the combats in the training set. The best combat model in out experiments is *TS-Lanchester*$^2$ with *static* DPF and a Borda Count target selection policy. While practically there is not a difference between the performance of our proposed models (in their best configuration, which is *static* DPF and *Borda Count* target selection); our proposed models can predict the final state significantly more accurately than SPARCRAFT. Additionally, our models can predict the final state with much lower computational requirements, being between 19 to 145 times faster than SPARCRAFT (although we'd like to point out that SPARCRAFT was not designed for this purpose, but for performing game-tree search at the scale of individual combats). This is specially important when we want to use a combat model as a forward model in the context of MCTS, where combat simulations need to be executed as part of the playouts.

Moreover, we noticed that simulating combats between heterogeneous armies (armies with different types of units) are particularly hard to predict. This is shown in Figure 5 where the final state similarity of each model is grouped by different types of combats: between homogeneous armies (1vs1), semi-homogeneous armies (1vsN) and heterogeneous armies (NvsN) (notice that "1vs1" does not mean "one unit versus one unit", but "an army composed of all units of one type versus another army composed of units of also one type"). As expected, there is a significant difference between heterogeneous armies and the rest; while the performance with semi-homogeneous armies is slightly better than homogeneous armies due the fact that normally there is a stronger army (the one with more than one type of unit).

Figure 6 shows the impact on the performance of the target selection policy in combats with heterogeneous armies (NvsN). We can observe how the learned Borda Count target policy outperforms the baseline (random) and a simple heuristic (destroy score).

### D. MCTS Abstraction Configuration

This section evaluates the performance of MCTS using each of the different abstractions presented in Section VII-A. To perform this evaluation, we incorporated our MCTS approach into the STARCRAFT bot Nova [28] and evaluated the performance of using MCTSCD [8] to command the army during a real game (all the military units are controlled by a single MCTS algorithm simulating the whole game).

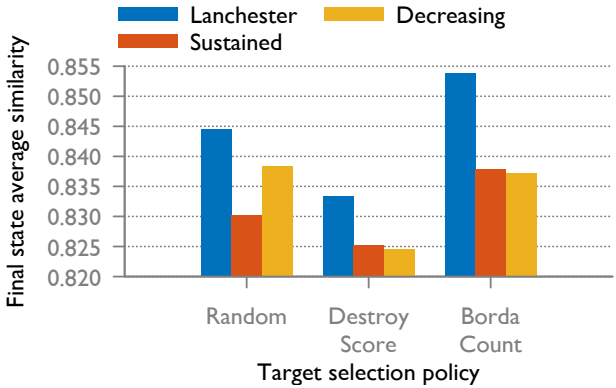

Fig. 6. Final state average similarity of each model in combats with heterogeneous armies (NvsN) clustered by target selection policy.

We deactivated *fog of war*, since dealing with partial-observability is out of scope in this paper. We also limited the length of a game to 20 minutes (28,800 frames), a reasonable limit since, in the STARCRAFT AI competition the average game length is about 21,600 frames (15 minutes), and usually the resources of the initial base are gone after 26,000 frames (18 minutes). MCTS is called once every 400 frames, and we paused the game while the search is taking place for experimentation purposes. As part of our future work, we want to explore splitting the search along several game frames, instead of pausing the game.

For our experiments we used the following MCTS configuration: $\epsilon$-greedy with $\epsilon = 0.2$ was used as the *tree policy*; the maximum *tree policy depth* set to 10; we used a random move selection for the *default policy*; the maximum *playout length* set to 2,880 game frames; for handling *simultaneous nodes* we use an *Alt* policy to decide which player will play; the computation budget was set to 10,000 playouts; and the *Sustain* model as a forward model. At the end of each playout, the following evaluation function was used to assess the game state:

$$score(A) = \sum_{1}^{n}(a_i.size \times a_i.destroyScore)$$

$$eval(A, B) = \frac{2 \times score(A)}{score(A) + score(B)} - 1$$

We used the STARCRAFT tournament map Benzene for our evaluation and we ran 100 games with our bot playing the Terran race against the built-in Terran AI of STARCRAFT.

For each abstraction we collected the average evaluation at the end of the game, the percentage of games won (without reaching the timeout), the percentage of lost games and the average length (in frames) of the games won. In Table IV we can observe how there is a large difference between using only regions or regions+chokepoints, this might be due to the increment on the search space that adding new nodes (chokepoints) generates. Then, between adding or not all the buildings (*-MA vs *-MB), adding all (*-MA) seems to help win more often and sooner. A thorough inspection of the evolution of the games over time, showed that the reason is because if only the bases are added, despite it being able to kill the enemy bases in less time, it did not kill the rest of the buildings (and hence win the game) because the game tree search was not able to "see" the remaining enemy buildings. Moreover, notice that even if the win percentage seems low (e.g., 60%), the R-MA configuration didn't lose any game, and in the 40% of games that timed out, our system was winning, but was just unable to find some of the last enemy units to finish the game off.

TABLE IV
EXPERIMENTS WITH DIFFERENT GAME STATE ABSTRACTIONS USING THE *Sustained* COMBAT MODEL.

| Abstraction | Avg. Eval | Win % | Loss % | Avg. Length |
|---|---|---|---|---|
| R-MA | **0.9391** | **60.0** | **0.0** | **24 870.7** |
| R-MB | 0.8604 | 25.0 | 4.0 | 25 924.2 |
| RC-MA | 0.5959 | 2.0 | 1.0 | 26 212.5 |
| RC-MB | 0.6139 | 2.0 | 1.0 | 27 561.0 |

TABLE V
EXPERIMENTS WITH DIFFERENT BOT CONFIGURATIONS

| Configuration | Avg. Eval | Win % | Loss % | Avg. Length |
|---|---|---|---|---|
| Scripted | 0.9683 | 98.0 | 2.0 | 18 542.5 |
| Random | 0.4403 | 3.0 | 3.0 | 25 463.3 |
| *TS-Lanchester*[2] | 0.9489 | 68.0 | 1.0 | **24 580.0** |
| *Sustained* | 0.9391 | 60.0 | **0.0** | 24 870.7 |
| *Decreasing* | **0.9682** | **75.0** | **0.0** | 24 732.1 |

### E. MCTS and Combat Model Performance

Finally, this section presents the performance achieved by MCTS using the different combat models proposed in this paper. We compare the performance against two baselines: the original version of the Nova bot (which uses a hard-coded manually tuned strategy for controlling the military units), and a *random* approach that issues random orders (at the same level of abstraction as the MCTS implementation). All the experiments were played against the built-in Terran AI. Unfortunately we could not test the MCTS with the combat models against other state-of-the-art STARCRAFT AIs due the limitation of no fog of war, and because we pause the game during MCTS search. The configuration for the experiment is the same as the previous sections with a R-MA abstraction.

In this experiment we collected the same information as the previous one. Results are shown in Table V. The first thing we see is that the *TS-Lanchester*[2] and *Decreasing* models both outperform the *Sustained* model, with the *Decreasing* model achieving the best performance (winning 75% of the games, and not losing any). Comparing these results against the two baselines, we can see that the *random* approach only manages to win 3% of the games, while losing another 3% (notice that this approach does not lose more often, since the underlying bot is still controlling unit production and micro-managing units during combat). On the other end, the *Scripted* approach manages to win 98% of the games, but loses 2% (while our MCTS approach using either *TS-Lanchester$^2$* or *Decreasing* lost any game). As a matter of fact, we can see that the average evaluation at the end of the games for *Decreasing* and *Scripted* is undistinguishable.

## X. CONCLUSIONS

This paper focused on the problem of deploying MCTS approaches in domains for which forward models are not available. Specifically, we focused on RTS games, and presented three combat models that can be used as combat forward models in RTS games. We also presented a method for training the parameters of these models from replay data. We have seen that in domains where the parameters of the models (damage per frame, target selection) are available from the game definition, those can be used directly. But in domains where this information is not available, it can be estimated from replay data.

Our results show that our combat models achieve better performance and are much faster than hand-crafted low-level models such as SPARCRAFT. This makes our proposed models suitable for MCTS approaches that need to perform a large number of simulations. All the combats models are very sensitive to the target selection policy, but we showed how the target selection can be learned from replay data. Finally, we show that the choice of combat model has a strong impact on the performance of MCTS, and our experiments in STARCRAFT showed that our *Decresing* model achieved the best performance.

As part of our future work we would like to explore online learning techniques to adapt the combat model to the current adversary, and also consider uncertainty and partial observability to be able to play with the fog of war activated. On the practical side, we would like to experiment with splitting the computation time spend in MCTS along several game frames, instead of pausing the game, in order to test our approach against other bots in the STARCRAFT competition.

**Alberto Uriarte** received the B.S. degree in Computer Science from Autonomous University of Barcelona (UAB), Spain, in 2006 and the M.S. degree in Computer Vision and Artificial Intelligence from Autonomous University of Barcelona (UAB), Spain, in 2011. He is currently pursuing the Ph.D. degree in computer science at Drexel University. His research interest includes game AI, RTS games, multiagents systems, procedural content generation, computational geometry and machine learning.

**Santiago Ontañón** is an assistant professor in the Computer Science Department at Drexel University. His main research interests are game AI, case-based reasoning and machine learning, fields in which he has published more than 100 peer-reviewed papers. He obtained his PhD form the Autonomous University of Barcelona (UAB), Spain. Before joining Drexel University, he held postdoctoral research positions at the Artificial Intelligence Research Institute (IIIA) in Barcelona, Spain, at the Georgia Institute of Technology (GeorgiaTech) in Atlanta, USA, and lectured at the University of Barcelona (UB), Spain.